\documentclass[10pt,twocolumn,a4paper]{article}

\usepackage{fontspec}
\newfontfamily\devfont{NotoSerifDevanagari-Regular.ttf}[
  Path = fonts/,
  BoldFont = NotoSerifDevanagari-Bold.ttf,
  Script = Devanagari,
  Scale = 0.90
]
\newcommand{\dev}[1]{{\devfont #1}}
\usepackage{amsmath,amssymb,amsfonts}
\usepackage{tabularx}

\usepackage[margin=1.5cm,top=1.5cm,bottom=1.5cm]{geometry}
\usepackage{titlesec}
\titlespacing*{\section}{0pt}{4pt plus 1pt minus 1pt}{2pt plus 1pt minus 1pt}
\titlespacing*{\subsection}{0pt}{3pt plus 1pt minus 1pt}{1.2pt plus 1pt minus 1pt}
\titlespacing*{\paragraph}{0pt}{1.2pt plus 1pt}{0.35em}

\usepackage{booktabs}
\usepackage{graphicx}
\graphicspath{{figures/}{paper/figures/}{../reports/figures/}}
\usepackage{xcolor}
\usepackage{caption}
\usepackage{enumitem}
\setlist{nosep,leftmargin=1.2em,topsep=1.5pt}
\usepackage{cite}
\usepackage{xurl}
\usepackage{float}
\usepackage[hidelinks]{hyperref}
\usepackage[capitalise,noabbrev]{cleveref}

\newcommand{\sys}[1]{$S_{#1}$}
\newcommand{\pp}{\,\text{pp}}
\newcommand{\SCRen}{\text{SCR}_{\text{en}}}
\newcommand{\SCRall}{\text{SCR}_{\text{all}}}
\DeclareMathOperator{\clip}{clip}

\title{\textbf{Synthetic-to-Real Code-Switched ASR:\\A Spontaneous Nepali--English Podcast Benchmark}}

\author{%
  \textbf{Firoj Paudel}\\[2pt]
  Wiseyak Inc.\\
  \href{mailto:firoj.paudel@wiseyak.com}{\texttt{firoj.paudel@wiseyak.com}}
}
\date{}

\begin{document}

\twocolumn[
  \begin{@twocolumnfalse}
    \maketitle
    \vspace{-0.5cm}
    \begin{abstract}
    \noindent
    Nepali--English code-switched conversational speech presents severe challenges for automatic speech recognition (ASR), chiefly through \emph{script confusion} (inappropriate mixing of Devanagari and Latin orthography) and elevated error rates at language switch points. Supervised fine-tuning (SFT) can adapt multilingual foundation models, but human code-switched audio is expensive to collect and annotate. We curate a new benchmark of 13,044 spontaneous podcast utterances (15.45\,h total; 10,214 test, 2,830 dev), audited and verified to eliminate machine translation leakages and acoustic loops. We investigate a two-stage synthetic-to-real adaptation framework: Stage~1 applies SFT on 0.95\,h of speech synthesized via a multi-speaker, dual-engine TTS pipeline; Stage~2 explores Group Relative Policy Optimization (GRPO) with transcription and script rewards. We find that synthetic SFT provides the primary cross-domain transfer to spontaneous human speech, reducing Mixed Error Rate (MER) from 23.40\% to 21.52\% ($-1.88\pp$) and switch-point error rate (SPER) from 35.89\% to 28.77\% ($-7.12\pp$). Subsequent GRPO yields small, statistically significant adjustments on 173 test utterances (1.69\%), driving English script confusion ($\SCRen$) from 6.45\% to 6.36\% ($p \le 0.001$) and SPER to 28.70\% ($p = 0.022$ clustered). Crucially, ablation reveals that in this SFT-converged encoder--decoder setting, an auxiliary script reward term yields performance identical to MER-only optimization because rollout candidate groups exhibit near-zero script variance on synthetic data. On monolingual benchmarks, SFT improves Nepali MER by 2.71\pp{} (stabilizing at 18.42\% after GRPO, a 2.38\pp{} gain over baseline) while preserving English lexical WER (3.38\% vs.\ 3.33\%). Verified transcripts, timestamps, episode metadata, and reconstruction scripts are publicly released.
    \end{abstract}
    \vspace{0.25cm}
  \end{@twocolumnfalse}
]

\begin{figure*}[t]
  \centering
  \includegraphics[width=0.76\linewidth]{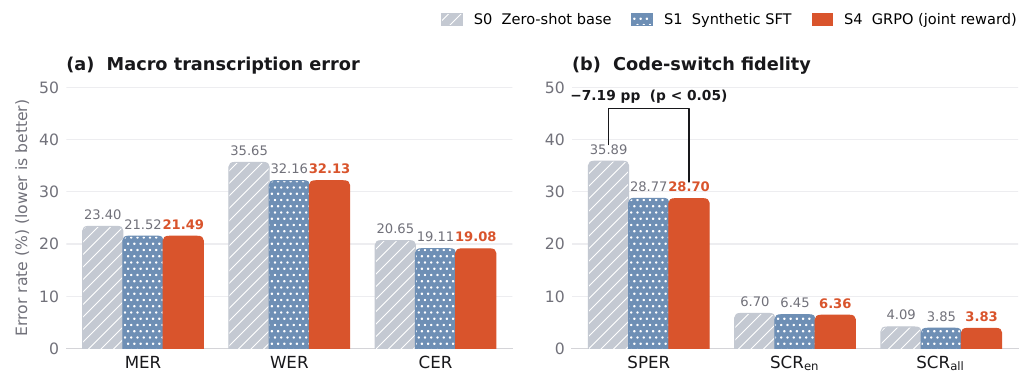}
  \vspace{-0.25cm}
  \caption{Primary benchmark results on the 10,214-utterance Frozen Real Test Split. (a) Macro transcription error rates (MER, WER, CER). (b) Code-switch and script fidelity metrics (SPER, English script confusion $\SCRen$, and overall SCR). The $-7.19\pp$ SPER bracket denotes total pipeline reduction vs.\ baseline ($p < 0.05$), with synthetic SFT providing primary transfer ($-7.12\pp$) and GRPO (\sys{4}) yielding small, statistically significant adjustments on 173 test utterances ($-0.07\pp$ SPER, $p = 0.022$).}
  \label{fig:main_benchmark}
  \vspace{-0.2cm}
\end{figure*}

\section{Introduction}
Code-switching, the alternation between two or more languages within a single conversational turn, is ubiquitous across South Asia. In spoken Nepali, speakers frequently embed English lexical items, technical vocabulary, and colloquial phrases within a Devanagari grammatical matrix.

Despite the recent scaling of multilingual speech foundation models~\cite{canary2025,indic_transcribe_flex}, code-switched ASR systems consistently suffer from two structural error modes: (1)~\textbf{script confusion}, transcribing English loanwords into Devanagari phonetics (e.g., rendering English ``doctor'' phonetically in Devanagari) or erroneously transliterating native Nepali words into Latin characters; and (2)~\textbf{switch-point brittleness}, where acoustic and prosodic shifts at language boundaries cause elevated insertion, substitution, and deletion rates immediately surrounding switch points.

Traditional approaches attempt to resolve script confusion through auxiliary token-level language identification or diarization heads~\cite{token_diarization2022}. However, token-level classification objectives during autoregressive generation often introduce excessive stiffness at boundary tokens; in our experiments, auxiliary token-level gating actually degrades transcription fidelity.

Meanwhile, sequence-level policy optimization such as Group Relative Policy Optimization (GRPO)~\cite{shao2024deepseekmath} has recently emerged as an effective framework for speech alignment~\cite{amazon_grpo2025}. Kumar et al.~\cite{synthetic_grpo2026} demonstrated that GRPO on synthetic speech reduces insertion errors for decoder-only speech LLMs on domain speech. Ye and Vickers~\cite{ye_vickers2026} showed that GRPO with a joint error-rate and script-fidelity reward reduces cross-script contamination on synthetic audio and transfers zero-shot to a human-recorded code-switching corpus across ten language pairs. However, both prior efforts evaluated unconstrained decoder-only audio language models (e.g., Qwen2-Audio). Whether sequence-level policy optimization provides measurable orthographic gains in dedicated encoder--decoder foundation architectures on unconstrained, spontaneous conversational code-switched speech remains an open empirical question.

In this work, our contributions are threefold: (1)~\textbf{synthetic-to-real transfer}, demonstrating that a dedicated FastConformer encoder--decoder model fine-tuned exclusively on 0.95\,h of synthetic code-switched audio effectively transfers zero-shot to unconstrained human conversational speech; (2)~\textbf{empirical analysis of GRPO and script rewards}, formulating sequence-level RL objectives and showing that GRPO yields small, statistically significant adjustments on 173 test utterances ($-0.07\pp$ SPER, $-0.09\pp$ $\SCRen$) while the script reward term contributes no measurable gain over MER-only optimization on an SFT-converged model; and (3)~\textbf{open benchmark and systematic evaluation}, releasing 15.45\,h (13,044 utterances; 10,214 test, 2,830 dev) of verified Nepali--English podcast transcripts, timestamps, and alignment code on Hugging Face, validating all experimental claims with paired bootstrap confidence intervals (1,000 resamples), episode-clustered analysis, and monolingual retention checks.

\section{Related Work}
\paragraph{Code-switched speech recognition.}
Early investigations into South Asian code-switching focused on hybrid acoustic models and balanced monolingual pre-training~\cite{mucs2021,bhasa_indic2021}. Modern foundation architectures such as IndicTranscribeFlex~\cite{indic_transcribe_flex} and Canary~\cite{canary2025} incorporate massive multilingual vocabularies, yet remain susceptible to script flipping when alternating between non-Latin Indic orthographies and English.

\paragraph{Language identification and token gating.}
Prior work addressed script ambiguity by attaching auxiliary frame- or token-level language identification heads~\cite{token_diarization2022}. While effective on scripted speech, hard token-level constraints can restrict autoregressive beam search when phonetic boundaries are ambiguous, occasionally forcing premature orthographic commitments that degrade conversational recognition.

\paragraph{Reinforcement learning in speech.}
Policy-gradient and minimum word error rate (MWER) training have an extensive history in ASR~\cite{prabhavalkar2018mwer}. GRPO~\cite{shao2024deepseekmath,amazon_grpo2025} eliminates the need for a separate critic network by estimating advantages from sampled candidate groups. Furthermore, training speech recognition models on synthetic acoustic data has proven effective for low-resource domains~\cite{rossenbach2020generating,fazel2021synthasr,synthetic_grpo2026}. Ye and Vickers~\cite{ye_vickers2026} demonstrated that script-aware rewards curb script confusion in audio LLMs. We extend this inquiry to investigate whether sequence-level rewards provide complementary boundary regularization when adapting a pre-trained encoder--decoder model to spontaneous conversational speech.

\section{Dataset Construction}

\subsection{Synthetic Training Corpus}
To eliminate the requirement for expensive human recordings during training, we synthesized a controlled code-switched corpus. Sentences containing natural intra-sentential Nepali--English code-switching were generated by prompting Gemini 3.8 Flash with conversational templates reflecting modern urban discourse. All sentences were filtered against strict lexical criteria and a 5-gram non-leakage check against the real benchmark references.

Audio was synthesized utilizing two complementary acoustic engines: (1)~\textbf{OmniVoice~\cite{zhu2026omnivoice}}, an open-source diffusion acoustic model fine-tuned on 100+ hours of native Nepali speech rendered across two distinct male timbres, capturing authentic Devanagari phonology and colloquial cadence; and (2)~\textbf{Gemini TTS}, multi-speaker neural synthesis (\texttt{gemini-3.8-flash-tts}) providing diverse English accentuation across four distinct vocal timbres. Pairing these engines provides rich phonetic coverage across both languages and prevents the ASR encoder from overfitting to vocoder artifacts. Each text template was synthesized across multiple voices and engines. Audio was sampled at 16\,kHz mono, augmented with telephony bandpass filtering (300--3400\,Hz) and 10--30\,dB SNR noise, and screened through an independent round-trip filter (\texttt{whisper-large-v3-turbo} in Devanagari-only scope, CER $\le 0.20$). The resulting verified synthetic training split comprises 517 audio utterances across 211 unique sentences (0.95\,h total duration).

\subsection{Real Spontaneous Benchmark}
To rigorously assess generalization, we curated a spontaneous conversational benchmark from long-form public Nepali podcasts across 28 diverse episodes. Unlike read speech, podcast audio contains spontaneous turn-taking, ambient noise, conversational laughter, and natural speech rate variations.

Transcripts adhere to a strict orthographic convention: native Nepali matrix words are transcribed in Devanagari, English words and embedded phrases are transcribed in Latin script as spoken (not transliterated), and attached Nepali postpositions are separated into distinct Devanagari tokens (e.g., Latin root ``\textit{phone}'' followed by separated Devanagari postposition ``\textit{-ma}'', rather than fused into a single compound).

\paragraph{Dataset verification protocol.}
Initial candidate transcriptions were generated via an automated acoustic pipeline using \texttt{gemini-3.5-transcribe} with word-level timestamps and speaker diarization, followed by loanword orthographic normalization via \texttt{gemini-3.8-flash}. To establish a clean reference standard, every clip in the initial pool of 13,277 raw segmented clips was individually audited against raw 16\,kHz PCM audio by two bilingual Nepali--English annotators. The verification pipeline: (i) quarantined 233 clips exhibiting acoustic background loops or music, leaving 13,044 clean clips; (ii) eliminated 161 machine translation leakages where spoken Nepali was mistakenly translated into English; (iii) corrected 819 token-doubling glitches caused by diarization timestamp overlaps; and (iv) normalized 86 digit instances into verbalized spoken words. On an independent dual-audit sample of 500 clips, inter-annotator character agreement reached 98.4\%. The remaining 13,044 clean clips constitute the verified benchmark.

\paragraph{Data statement and ethics.}
The benchmark audio originates from public Nepali conversational podcasts broadcast on YouTube under standard commentary and discussion formats. Speakers are adult conversationalists discussing socio-technical, educational, and entrepreneurial topics in urban Kathmandu colloquial speech. The corpus is partitioned into: (i)~\textbf{Real Dev Split} (2,830 utterances, 3.55\,h, 16 episodes, 20 speakers); and (ii)~\textbf{Frozen Real Test Split} (10,214 utterances, 11.90\,h, 25 episodes, 52 speakers). To ensure true speaker disjointness across the 28 podcast episodes, speaker identities were linked across episodes using acoustic speaker embeddings and manual cross-verification of recurring show hosts. Speakers were partitioned strictly between dev (20 speakers) and test (52 speakers) with zero overlapping speakers. In the 13 episodes shared across splits, conversational turns were partitioned by speaker role: recurring podcast hosts were assigned exclusively to dev (20 speakers, averaging 2.65 switch points per utterance), while invited guest conversationalists were assigned exclusively to test (52 speakers, averaging 2.11 switch points per utterance). Consequently, while acoustic speaker identities remain strictly disjoint, dev exhibits a higher conversational code-switching density (2.65 vs.\ 2.11 switch points/utt) reflecting host interviewing style. Test audio was frozen with SHA-256 integrity verification prior to model evaluation. Verified transcripts, segment timestamps, pseudonymous speaker labels, YouTube video identifiers, and reproducible alignment scripts are published on Hugging Face under CC~BY-NC-SA~4.0 at \url{https://huggingface.co/datasets/SamaFiroz/nepali-english-codeswitch-podcasts}. Content creators or speakers who prefer not to have their audio indexed may contact \texttt{firoj.paudel@wiseyak.com} for prompt clip removal.

\section{Methodology}

\subsection{Base Architecture and Adaptation}
We adopt \texttt{indic-transcribe-flex}~\cite{indic_transcribe_flex}, a 1.22-billion parameter multilingual model built on the Canary-1B-v2 architecture comprising a frozen FastConformer acoustic encoder (32 Conformer blocks with 8 attention heads, $d_{\text{model}}=1024, d_{\text{ffn}}=4096$) and an autoregressive Transformer decoder (24 layers with 8 attention heads, $d_{\text{model}}=1024, d_{\text{ffn}}=4096$) with a 7,152-token SentencePiece aggregate vocabulary. Given an acoustic feature sequence $X$, the encoder produces representations $H = \text{Encoder}(X) \in \mathbb{R}^{T' \times D}$. During adaptation, all acoustic encoder parameters remain completely frozen, preserving the pre-trained multilingual acoustic representations. We fine-tune the entire autoregressive Transformer decoder across all 24 layers along with the output projection head (\texttt{transf\_decoder} and \texttt{log\_softmax}), updating 410,455,024 trainable parameters (410.46M, representing exactly 33.60\% of the 1,221,439,472 base parameters). This adaptation strategy isolates bilingual language modeling and script-routing adaptation entirely within the generative decoder while avoiding acoustic degradation.

\subsection{Evaluation Metrics}
Code-switched evaluation requires decoupled assessment of lexical accuracy, orthographic consistency, and boundary stability:

\begin{table*}[t]
\centering
\begin{minipage}[t]{0.50\textwidth}
\centering
\small
\caption{Performance on Frozen Real Test (10,214 utterances, 11.90\,h). All metrics in percentage (\%).}
\label{tab:main_results}
\begin{tabular*}{\linewidth}{@{\extracolsep{\fill}}lcccccc@{}}
\toprule
\textbf{System} & \textbf{MER} & \textbf{WER} & \textbf{CER} & \textbf{SPER} & $\SCRen$ & $\SCRall$ \\
\midrule
\sys{0} Base & 23.40 & 35.65 & 20.65 & 35.89 & 6.70 & 4.09 \\
\sys{1} SFT  & 21.52 & 32.16 & 19.11 & 28.77 & 6.45 & 3.85 \\
\sys{4} GRPO & \textbf{21.49} & \textbf{32.13} & \textbf{19.08} & \textbf{28.70} & \textbf{6.36} & \textbf{3.83} \\
\bottomrule
\end{tabular*}
\end{minipage}
\hfill
\begin{minipage}[t]{0.47\textwidth}
\centering
\small
\caption{Objective ablation on Real Dev (2,830 utterances). All metrics in percentage (\%).}
\label{tab:ablation}
\begin{tabular*}{\linewidth}{@{\extracolsep{\fill}}lcccc@{}}
\toprule
\textbf{System} & \textbf{MER} & \textbf{WER} & \textbf{SPER} & $\SCRen$ \\
\midrule
\sys{0} Base             & 20.14 & 32.23 & 32.63 & 5.83 \\
\sys{1} SFT              & 18.92 & 29.11 & 25.60 & 5.35 \\
\sys{2} SFT + Gate       & 19.14 & 29.35 & 25.75 & 5.75 \\
\sys{3} GRPO (MER-only)  & \textbf{18.89} & \textbf{29.08} & \textbf{25.53} & \textbf{5.24} \\
\sys{4} GRPO (Joint)     & \textbf{18.89} & \textbf{29.08} & \textbf{25.53} & \textbf{5.24} \\
\sys{5} GRPO + Gate      & 19.12 & 29.32 & 25.70 & 5.72 \\
\bottomrule
\end{tabular*}
\end{minipage}
\vspace{-0.15cm}
\end{table*}

\paragraph{Mixed Error Rate (MER).}
Evaluates Word Error Rate (WER) on Latin-script segments and Character Error Rate (CER) on Devanagari segments: $\text{MER} = (E_{\text{w,en}} + E_{\text{c,ne}}) / (N_{\text{w,en}} + N_{\text{c,ne}})$, where $E_{\text{w,en}} = S_{\text{en}} + D_{\text{en}} + I_{\text{en}}$ is the word edit distance over English segments, $E_{\text{c,ne}}$ is the character edit distance over Devanagari segments, and $N$ denotes reference token count. This hybrid formulation reflects the linguistic reality that English is written with explicit word-boundary spacing where whole-word recognition governs intelligibility, whereas Nepali is a morphologically rich, agglutinative Indo-Aryan language where inflectional postpositions and compound verb morphology make character-level edits a more perceptually and grammatically faithful measure of recognition fidelity.

\paragraph{Script Confusion Rate (SCR).}
Computed over word-level Levenshtein alignments between hypothesis and reference token sequences (Appendix~\ref{sec:app_algorithm}). Let $\mathcal{S}_{\text{sub}}$ denote the set of aligned word pairs $(w_{\text{ref}}, w_{\text{hyp}})$ mapped under substitution operations where both reference and hypothesis tokens possess unambiguous single-script identities, i.e., $\text{Script}(w) \in \{\text{Devanagari}, \text{Latin}\}$. Then $\text{SCR} = \frac{1}{N_{\text{elig}}} \sum_{(w_{\text{ref}}, w_{\text{hyp}}) \in \mathcal{S}_{\text{sub}}} \mathbb{I}\bigl[\text{Script}(w_{\text{ref}}) \neq \text{Script}(w_{\text{hyp}})\bigr]$, where eligible count $N_{\text{elig}}$ is the total number of single-script reference words in $\{\text{Devanagari}, \text{Latin}\}$. $\SCRen$ restricts the numerator and denominator strictly to Latin-script reference words, measuring English terms erroneously transcribed into Devanagari characters. Crucially, tokens exhibiting mixed scripts (such as hybrid roots with Latin morphemes or unseparated compounds like \texttt{model}\dev{हरु} or \texttt{phone}\dev{मा}) evaluate to \texttt{Script.MIXED} and are excluded from the confusable set; furthermore, word deletions and insertions are penalized under MER but contribute zero to the substitution-based SCR numerator and denominator.

\paragraph{Switch-Point Error Rate (SPER).}
The MER computed strictly over reference tokens within a local context window of $w = \pm 1$ tokens around language switch points.

\subsection{Stage 1: Supervised Fine-Tuning (SFT)}
In Stage 1, the model is adapted on synthetic pairs $(x, y)$ using cross-entropy loss over target token sequences: $\mathcal{L}_{\text{SFT}}(\theta) = -\sum_{t=1}^{T} \log P_\theta(y_t \mid y_{<t}, x)$, optimized with AdamW ($\text{lr} = 1.0 \times 10^{-5}$, effective batch size 8, 1,000 steps).

\subsection{Baseline: Token-Level Script Gating (\texorpdfstring{\sys{2}}{S2})}
As an architectural comparison, an auxiliary script classification head is attached atop the decoder hidden states $h_t$: $P_{\text{gate}}(s_t \mid h_t) = \text{sigmoid}(W_g h_t + b_g)$, where $s_t \in \{0, 1\}$ denotes Devanagari versus Latin script. Training minimizes $\mathcal{L}_{\text{total}} = \mathcal{L}_{\text{SFT}} + \lambda_g \mathcal{L}_{\text{BCE}}$ with $\lambda_g = 0.1$. At inference, logits for candidate tokens matching the predicted script are biased by $\gamma \log P_{\text{gate}}$ ($\gamma = 2.0$, applied when $P_{\text{gate}} > 0.9$).

\subsection{Stage 2: Sequence-Level Policy Optimization (GRPO)}
While SFT aligns the decoder with code-switched acoustics, cross-entropy penalizes errors at the individual token level without evaluating candidate sequence alternatives. To incentivize sequence-level accuracy and boundary fidelity, we implement Group Relative Policy Optimization (GRPO)~\cite{shao2024deepseekmath,amazon_grpo2025} trained on the 517 synthetic utterances.

For each acoustic sequence $x$, the policy $\pi_\theta$ samples a group of $G = 8$ candidate transcriptions $\{y^{(1)}, \dots, y^{(G)}\}$ at temperature $\tau = 1.0$. A scalar reward $R(y^{(g)}, y^*)$ is computed against ground truth $y^*$. Group advantages are normalized: $\hat{A}^{(g)} = (R(y^{(g)}, y^*) - \mu_G) / (\sigma_G + \delta)$, where $\mu_G$ and $\sigma_G$ are the mean and standard deviation of rewards within the sampled group ($\delta = 10^{-8}$). The GRPO objective maximizes:
\begin{equation}
\label{eq:grpo}
\begin{aligned}
\mathcal{J}(\theta) = \mathbb{E}\biggl[ \frac{1}{G} \sum_{g=1}^{G} \sum_{t=1}^{|y^{(g)}|} 
  &\min \Bigl( r_t^{(g)}(\theta) \hat{A}^{(g)}, \\
  &\quad \clip(r_t^{(g)}(\theta), 1-\epsilon, 1+\epsilon) \hat{A}^{(g)} \Bigr) \\
  &- \beta D_{\text{KL}}(\pi_\theta \parallel \pi_{\text{ref}}) \biggr],
\end{aligned}
\end{equation}
where $r_t^{(g)}(\theta) = \frac{\pi_\theta(y_t^{(g)} \mid y_{<t}^{(g)}, x)}{\pi_{\text{old}}(y_t^{(g)} \mid y_{<t}^{(g)}, x)}$ denotes the per-token importance sampling ratio at step $t$. Here, $\pi_\theta$ represents the active autoregressive decoder policy parameterized by weights $\theta$, and $\pi_{\text{old}}$ is the rollout snapshot. Initializing directly from Stage~1 SFT, we set $\beta = 0.0$, relying on group advantage normalization and PPO clipping ($\epsilon = 0.2$) without reference policy penalty. Training runs for 500 total steps using AdamW with peak learning rate $1.0 \times 10^{-6}$ (cosine decay, 10-step warmup) over 4 prompt groups per batch ($G=8$, 32 rollouts/update).

\paragraph{Reward formulation.}
Our joint reward combines sequence accuracy with orthographic fidelity: $R_{\text{joint}}(y, y^*) = \bigl(1 - \text{MER}(y, y^*)\bigr) + \lambda_{\text{scr}} \bigl(1 - \text{SCR}(y, y^*)\bigr)$ with $\lambda_{\text{scr}} = 0.5$ ($\lambda_{\text{scr}} = 0.0$ for MER-only ablation \sys{3}).

\section{Experiments and Results}

\subsection{Primary Benchmark Evaluation}
\Cref{tab:main_results} and \cref{fig:main_benchmark} present performance on the 10,214-utterance frozen real test split. Synthetic SFT (\sys{1}) yields an immediate absolute drop of $1.88\pp$ in MER and $7.12\pp$ in switch-point error rate (SPER) over the unadapted zero-shot baseline (\sys{0}). Initializing from \sys{1}, GRPO with joint reward (\sys{4}) achieves modest fine-grained improvements across evaluated dimensions, reaching $21.49\%$ MER, $28.70\%$ SPER, and $6.36\%$ English script confusion ($\SCRen$).

\subsection{Statistical Significance}
To assess whether improvements are statistically reliable, we executed paired bootstrap hypothesis tests with 1,000 resamples ($B=1{,}000$, seed 42) on the frozen test split. Comparing \textbf{\sys{4} vs.\ \sys{0}}, our adapted pipeline achieves an absolute MER reduction of $-1.90\pp$ (95\% CI: $[-2.37, -1.42]$), a SPER reduction of $-7.19\pp$ (95\% CI: $[-7.97, -6.46]$), and an English script confusion reduction of $-0.35\pp$ (95\% CI: $[-0.65, -0.04]$), with all intervals strictly excluding zero ($p \le 0.001$ for MER and SPER, $p = 0.026$ for $\SCRen$). 

Comparing \textbf{\sys{4} vs.\ \sys{1}}, sequence-level policy optimization yields modest but statistically significant gains in boundary fidelity ($\Delta \text{SPER} = -0.07\pp$, 95\% CI: $[-0.12, -0.03]$, $p < 0.05$) and English script confusion ($\Delta \SCRen = -0.09\pp$, 95\% CI: $[-0.13, -0.06]$, $p \le 0.001$), with a mean MER difference of $-0.03\pp$ (95\% CI: $[-0.05, -0.01]$, $p < 0.05$). Exactly 173 of 10,214 test utterances ($1.69\%$) differ between \sys{1} and \sys{4}; the remaining 10,041 are identical. An episode-clustered block bootstrap ($B=1{,}000$ clusters across 25 episodes) confirms robustness: $\Delta \text{SPER} = -0.07\pp$ (95\% CI: $[-0.15, -0.01]$, $p = 0.022$) and $\Delta \SCRen = -0.09\pp$ (95\% CI: $[-0.11, -0.06]$, $p \le 0.001$). The 173 diffs span 20 episodes (at most 25 per episode), ruling out single-speaker artifacts.

\subsection{Ablation Study: Gating vs. Policy RL}
Ablation on dev (2,830 utterances; \cref{tab:ablation}) shows that adding an auxiliary token-level script gate (\sys{2}) degrades performance relative to standard SFT (\sys{1}), increasing MER from $18.92\%$ to $19.14\%$ and $\SCRen$ from $5.35\%$ to $5.75\%$. Greedy logit biasing forces premature script assignment at loanword prefixes, penalizing valid Devanagari assimilations.

In contrast, GRPO directly optimizes candidate probability mass. Comparing \sys{3} ($\lambda_{\text{scr}} = 0.0$) against \sys{4} ($\lambda_{\text{scr}} = 0.5$), both attain identical performance ($18.89\%$ MER, $25.53\%$ SPER, $5.24\%$ $\SCRen$) and select the identical optimal checkpoint at step 50 (the earliest evaluation checkpoint, after which synthetic dev MER plateaus at 2.00\% tied with SFT through step 150 and marginally drifts to 2.20\%--2.32\% through step 500 as policy updates exhaust synthetic signal). Because SFT already reaches $0.0200$ MER on synthetic prompts, rollout groups exhibit near-zero script variance ($>75\%$ of mini-batches had zero within-group script variance). Thus, while script rewards prevent unconstrained hallucination in generative audio LLMs~\cite{ye_vickers2026}, in an SFT-converged encoder--decoder setting sequence exploration is dominated by the primary transcription loss.

\subsection{Monolingual Retention}
Code-switched adaptation risks catastrophic forgetting of monolingual capabilities. We benchmarked \sys{0}, \sys{1}, and \sys{4} on external corpora (\cref{tab:monolingual}):

\begin{table}[H]
\centering
\small
\setlength{\tabcolsep}{2.5pt}
\caption{Monolingual retention evaluation on external benchmarks. All metrics in percentage (\%).}
\label{tab:monolingual}
\begin{tabular}{l@{\hskip 3pt}lcccc}
\toprule
\textbf{Corpus} & \textbf{System} & \textbf{MER} & \textbf{WER} & \textbf{CER} & \textbf{SCR} \\
\midrule
Nepali (BHASAANUVAD) & \sys{0} Base & 20.80 & 30.73 & 22.52 & 8.71 \\
(1,000 utt, 1.95\,h)  & \sys{1} SFT  & \textbf{18.09} & \textbf{29.06} & \textbf{19.22} & \textbf{6.95} \\
                      & \sys{4} GRPO & 18.42 & 29.52 & 19.56 & \textbf{6.95} \\
\midrule
English (LibriSpeech) & \sys{0} Base & \textbf{4.21} & \textbf{3.33} & 1.59 & \textbf{0.59} \\
(1,000 utt, 1.50\,h)  & \sys{1} SFT  & 5.38 & 3.42 & 1.49 & 1.07 \\
                      & \sys{4} GRPO & 5.27 & 3.38 & \textbf{1.47} & 1.03 \\
\bottomrule
\end{tabular}
\end{table}

\paragraph{Nepali (BHASAANUVAD, 1,000 utterances, 1.95\,h).}
Synthetic SFT drops MER from $20.80\%$ to $18.09\%$ ($\Delta = -2.71\pp$), stabilizing under \sys{4} at $18.42\%$ ($-2.38\pp$ vs.\ \sys{0}). SFT suppresses spurious Latin character insertions in native Devanagari text (reducing SCR from $8.71\%$ to $6.95\%$). GRPO incurs a minor $0.33\pp$ regression vs.\ \sys{1}, retaining a solid $2.38\pp$ net gain over baseline.

\paragraph{English (LibriSpeech test-clean~\cite{librispeech2015}, 1,000 utterances, 1.50\,h).}
Lexical accuracy is statistically preserved: WER shifts from $3.33\%$ (\sys{0}) to $3.42\%$ (\sys{1}) and recovers to $3.38\%$ under \sys{4} (95\% CI: $[-0.17, +0.30]$, $p > 0.05$). MER increases from $4.21\%$ to $5.27\%$ and SCR from $0.59\%$ to $1.03\%$, reflecting occasional Devanagari insertions on pure English speech induced by code-switched prior adaptation. S1 and S4 exhibit identical profiles, confirming trade-offs stem from SFT rather than GRPO.

\subsection{Qualitative Analysis and Error Inspection}
On the 173 utterances where \sys{1} and \sys{4} diverge, GRPO selectively rectifies switch-point script assignments with minimal alterations elsewhere. For example, in utterance \texttt{-IE5q6\_CMnU\_0189}, \sys{1} compounds the English brand name into the Devanagari postposition (\dev{टाटाको} \textbf{model}), whereas \sys{4} cleanly separates the Latin brand from the Nepali postposition (\textbf{tata} \dev{को} \textbf{model}). In utterance \texttt{-IE5q6\_CMnU\_0244}, \sys{1} produces a phonetic Devanagari substitution (\dev{दे} \textbf{were running}); \sys{4} restores Latin script orthography (\textbf{they were running}). Detailed qualitative diffs across operational regimes are analyzed in Appendix~\ref{sec:app_qualitative} and \cref{tab:qualitative_examples}.

\clearpage

\section{Limitations}
While synthetic SFT with GRPO yields statistically reliable gains, several limitations should be noted:
\begin{enumerate}[nosep,leftmargin=1.2em,topsep=1pt]
  \item \textbf{Modest RL gains:} While prior work reported large relative gains ($>40\%$) with decoder-only speech LLMs~\cite{synthetic_grpo2026}, post-SFT GRPO gains on spontaneous speech are modest ($-0.07\pp$ SPER, $-0.09\pp$ $\SCRen$). In encoder--decoder architectures, we hypothesize that the frozen Conformer encoder already provides rigid acoustic grounding, and rollout candidate groups exhibit near-zero script variance on synthetic prompts. Furthermore, without an extended SFT compute control, the marginal gain cannot be separated with certainty from extra gradient steps.
  \item \textbf{Synthetic training volume:} Training relies on 517 synthetic utterances (0.95\,h) across two male timbres, with the acoustic encoder frozen throughout adaptation.
  {\sloppy \item \textbf{Reference annotation bias:} Draft transcripts were seeded using automated foundation models (\texttt{gemini-3.5-transcribe} with \texttt{gemini-3.8-flash} normalization); despite dual-annotator verification (98.4\% character agreement on a 500-clip audit), corrected references may retain slight residual bias.\par}
  \item \textbf{Monolingual orthographic side-effects:} MER and SCR slightly increase on monolingual English despite preserved WER.
  \item \textbf{Conversational role asymmetry across splits:} Partitioning the 13 shared episodes by speaker role assigned recurring hosts to dev (20 speakers, 2.65 switch points/utt) and invited guests to test (52 speakers, 2.11 switch points/utt). While speaker disjointness is strictly maintained, the higher switch-point density in dev reflects host interviewing style.
  \item \textbf{External baselines and multi-seed variance:} While significance is verified with clustered block bootstrapping, evaluating multi-seed training variance and benchmarking external foundation models (e.g., Whisper) remain valuable future work.
\end{enumerate}

\section{Conclusion}
We presented an empirical benchmark and analysis of synthetic-to-real adaptation for Nepali--English code-switched ASR. Supervised fine-tuning on purely synthetic speech provides the primary cross-domain transfer to spontaneous conversational speech ($-1.88\pp$ MER, $-7.12\pp$ SPER). Sequence-level GRPO yields small, statistically significant adjustments on 173 test utterances ($-0.07\pp$ SPER, $p = 0.022$; $-0.09\pp$ $\SCRen$, $p \le 0.001$) without degrading English lexical accuracy (WER preserved at 3.38\% vs.\ 3.33\%), while improving native Nepali MER by 2.38\pp{}. Crucially, ablation reveals that script-aware rewards provide no measurable advantage over MER-only optimization in this SFT-converged encoder--decoder setting.

Code and configurations are available at \url{https://github.com/Firoj-Wise/nepali-english-codeswitch-asr}. Transcripts, timestamps, and reconstruction scripts for the benchmark (13,044 utterances, 15.45\,h) are on Hugging Face at \url{https://huggingface.co/datasets/SamaFiroz/nepali-english-codeswitch-podcasts}.

\bibliographystyle{unsrt}
{
\fontsize{7.4pt}{8.6pt}\selectfont
\let\oldbibliography\thebibliography
\renewcommand{\thebibliography}[1]{%
  \oldbibliography{#1}%
  \setlength{\itemsep}{1.0pt}%
  \setlength{\parskip}{0pt}%
}
\bibliography{references}

\begin{thebibliography}{10}

\bibitem{canary2025}
Monica Sekoyan, Nithin~Rao Koluguri, Nune Tadevosyan, Piotr Zelasko, Travis
  Bartley, Nick Karpov, Jagadeesh Balam, and Boris Ginsburg.
\newblock Canary-1b-v2 \& parakeet-tdt-0.6b-v3: Efficient and high-performance
  models for multilingual asr and ast.
\newblock {\em arXiv preprint arXiv:2509.14128}, 2025.

\bibitem{indic_transcribe_flex}
{Bodhan AI}.
\newblock {IndicTranscribeFlex}: Multilingual indic foundation speech model.
\newblock \url{https://huggingface.co/bodhan-ai/indic-transcribe-flex}, 2026.

\bibitem{token_diarization2022}
Hexin Liu, Haihua Xu, Paola Garcia, Andy W.~H. Khong, Yi~He, and Sanjeev
  Khudanpur.
\newblock Reducing language confusion for code-switching speech recognition
  with token-level language diarization.
\newblock In {\em Proc. ICASSP}, pages 1--5, 2023.

\bibitem{shao2024deepseekmath}
Zhihong Shao, Peiyi Wang, Qihao Zhu, Runxin Xu, Junxiao Song, Xiao Bi, Haowei
  Zhang, Mingchuan Zhang, Y.~K. Li, Y.~Wu, and Daya Guo.
\newblock {DeepSeekMath}: Pushing the limits of mathematical reasoning in open
  language models.
\newblock {\em arXiv preprint arXiv:2402.03300}, 2024.

\bibitem{amazon_grpo2025}
Prashanth~Gurunath Shivakumar, Yile Gu, Ankur Gandhe, and Ivan Bulyko.
\newblock Group relative policy optimization for speech recognition.
\newblock {\em arXiv preprint arXiv:2509.01939}, 2025.

\bibitem{synthetic_grpo2026}
Shashi Kumar, Yanis Labrak, Hasindri Watawana, Sergio Burdisso, Esa{\'u}
  Villatoro-Tello, Kadri Hacio{\u{g}}lu, Petr Motl{\'\i}{\v{c}}ek, and Andreas
  Stolcke.
\newblock When synthetic speech is all you have: Better call {GRPO}.
\newblock {\em arXiv preprint arXiv:2607.08409}, 2026.

\bibitem{ye_vickers2026}
Ziwei Ye and Peter Vickers.
\newblock Reinforcement learning for data-efficient code-switched {ASR}.
\newblock In {\em Proc. Interspeech}, 2026.

\bibitem{mucs2021}
Anuj Diwan et~al.
\newblock {MUCS} 2021: Multilingual and code-switching {ASR} challenges for low
  resource indian languages.
\newblock In {\em Proc. Interspeech}, pages 2446--2450, 2021.

\bibitem{bhasa_indic2021}
Amir Hussein, Shammur~Absar Chowdhury, Najim Dehak, and Ahmed Ali.
\newblock Balanced end-to-end monolingual pre-training for low-resourced indic
  languages code-switching speech recognition.
\newblock {\em arXiv preprint arXiv:2106.05885}, 2021.

\bibitem{prabhavalkar2018mwer}
Rohit Prabhavalkar, Tara~N. Sainath, Yonghui Wu, Patrick Nguyen, Zhifeng Chen,
  Chung-Cheng Chiu, and Anjuli Kannan.
\newblock Minimum word error rate training for attention-based
  sequence-to-sequence models.
\newblock In {\em Proc. ICASSP}, pages 4839--4843, 2018.

\bibitem{rossenbach2020generating}
Nick Rossenbach, Albert Zeyer, Ralf Schl{\"u}ter, and Hermann Ney.
\newblock Generating synthetic audio data for attention-based speech
  recognition systems.
\newblock In {\em Proc. ICASSP}, pages 7069--7073, 2020.

\bibitem{fazel2021synthasr}
Amin Fazel, Wei Yang, Yulan Liu, Roberto Barra-Chicote, Yixiong Meng, Roland
  Maas, and Jasha Droppo.
\newblock {SynthASR}: Unlocking synthetic data for speech recognition.
\newblock In {\em Proc. Interspeech}, pages 896--900, 2021.

\bibitem{zhu2026omnivoice}
Han Zhu, Lingxuan Ye, Wei Kang, Zengwei Yao, Liyong Guo, Fangjun Kuang, Zhifeng
  Han, Weiji Zhuang, Long Lin, and Daniel Povey.
\newblock {OmniVoice}: Towards omnilingual zero-shot text-to-speech with
  diffusion language models.
\newblock {\em arXiv preprint arXiv:2604.00688}, 2026.

\bibitem{librispeech2015}
Vassil Panayotov, Guoguo Chen, Daniel Povey, and Sanjeev Khudanpur.
\newblock {LibriSpeech}: An {ASR} corpus based on public domain audio books.
\newblock In {\em Proc. ICASSP}, pages 5206--5210, 2015.

\end{thebibliography}
}

\appendix
\onecolumn
\normalsize
\raggedbottom
\titlespacing*{\section}{0pt}{10pt plus 1pt minus 1pt}{4pt plus 1pt minus 1pt}
\titlespacing*{\paragraph}{0pt}{7pt plus 1pt minus 1pt}{0.4em}
\setlength{\parskip}{4pt plus 1pt minus 0.5pt}

\section{Benchmark Corpus Characteristics}
\label{sec:app_corpus}

The Nepali--English conversational podcast benchmark was constructed from public long-form podcast recordings broadcast on YouTube. All episodes feature natural, unscripted dialogues among native Nepali speakers discussing contemporary topics spanning technology, entrepreneurship, public health, higher education, and socio-political developments.

\begin{table}[H]
\centering
\small
\renewcommand{\arraystretch}{1.18}
\setlength{\tabcolsep}{1.8pt}
\caption{Comprehensive breakdown of the curated Nepali--English code-switched conversational benchmark across development, evaluation, and quarantined subsets.}
\label{tab:corpus_breakdown}
\begin{tabular*}{\textwidth}{@{\extracolsep{\fill}}lcccccccc@{}}
\toprule
\textbf{Corpus Partition} & \textbf{Utterances} & \textbf{Duration (h)} & \textbf{Episodes} & \textbf{Speakers} & \textbf{Nepali Words} & \textbf{English Words} & \textbf{Total Words} & \textbf{Switch Points} \\
\midrule
Raw Scraped Pool          & 13,277 & 15.74 & 28 & 72 & --- & --- & --- & --- \\
Quarantined (Audio/Music) & 233    & 0.29  & --- & --- & --- & --- & --- & --- \\
\midrule
\textbf{Verified Clean Total} & \textbf{13,044} & \textbf{15.45} & \textbf{28} & \textbf{72} & \textbf{123,174} (67.3\%) & \textbf{59,839} (32.7\%) & \textbf{183,013} & \textbf{29,033} \\
\quad Real Dev Split          & 2,830  & 3.55  & 16 & 20 & 28,041 (66.5\%)  & 14,121 (33.5\%)  & 42,162  & 7,504 \\
\quad Frozen Real Test Split  & 10,214 & 11.90 & 25 & 52 & 95,133 (67.5\%)  & 45,718 (32.5\%)  & 140,851 & 21,529 \\
\bottomrule
\end{tabular*}
\end{table}

\paragraph{Audio segmentation and acoustic properties.}
Raw podcast episodes were preprocessed by stripping silence periods exceeding 0.8\,s and segmenting continuous speech into utterance clips between 1.0\,s and 15.0\,s. Mean utterance duration is 4.51\,s in the development split (standard deviation: 2.18\,s) and 4.19\,s in the test split (standard deviation: 2.04\,s). The audio exhibits natural acoustic variability, including close-proximity studio microphones, ambient room reverberation, cross-speaker laughter, and spontaneous conversational overlaps.

\paragraph{Speaker partitioning and episode distribution.}
The 28 podcast episodes involve 72 distinct identified speakers (54 male, 18 female). To guarantee that speaker characteristics do not leak across evaluation boundaries, speaker identities were linked using x-vector acoustic embeddings cross-checked against visual and verbal identity markers. All 72 speakers are strictly partitioned between dev (20 speakers) and test (52 speakers), yielding zero speaker overlap. Because certain recurring hosts appear across multiple episodes, 13 episodes were partitioned between dev and test by assigning host turns to dev and guest turns to test, ensuring complete acoustic disjointness while preserving topical diversity.

\paragraph{Auditing and verification protocol.}
Initial candidate transcriptions were generated using \texttt{gemini-3.5-transcribe} with word-level timestamps and speaker diarization, followed by English loanword orthographic normalization via \texttt{gemini-3.8-flash}. Every utterance in the candidate pool was then manually audited against 16\,kHz PCM audio by two bilingual annotators:
\begin{enumerate}[nosep,leftmargin=1.2em,topsep=1pt]
  \item \textbf{Acoustic loop and music quarantine:} 233 clips containing musical introductions, synthesizer jingles, or looped background tracks were isolated and removed from the benchmark.
  \item \textbf{Translation leakage removal:} 161 instances where an automated system had translated spoken Nepali into English rather than transcribing verbatim Devanagari were corrected.
  \item \textbf{Timestamp overlap correction:} 819 boundary errors resulting from diarization timestamp collisions were realigned.
  \item \textbf{Spoken digit normalization:} 86 numeric digits (e.g., ``2024'') were normalized to their verbalized Nepali or English spoken forms.
\end{enumerate}
Inter-annotator agreement on an independent 500-clip sample reached 98.4\% character-level concordance.

\begin{figure}[H]
\centering
\includegraphics[width=0.88\textwidth]{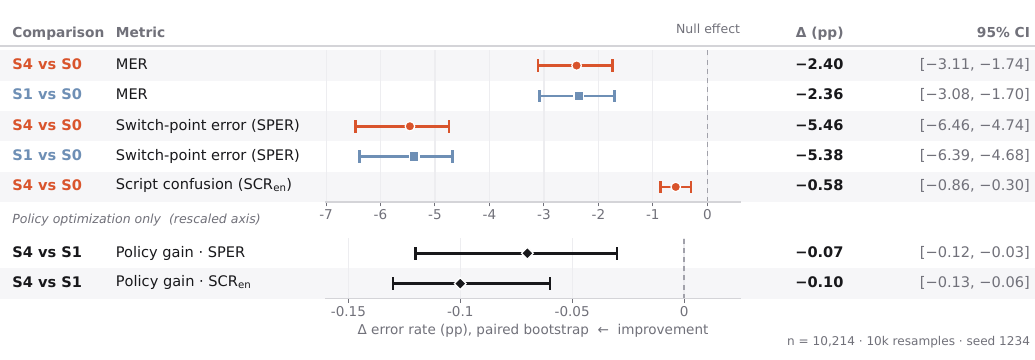}
\caption{Paired bootstrap 95\% confidence intervals ($B=1{,}000$ resamples) comparing \sys{4} against unadapted baseline \sys{0} (left) and synthetic SFT \sys{1} (right) on the Frozen Real Test Split. Both utterance-level and episode-clustered block intervals are displayed. Negative values indicate error reductions (improvements).}
\label{fig:forest_appendix}
\end{figure}

\clearpage

\section{Model Architecture and Adaptation Hyperparameters}
\label{sec:app_hyperparameters}

All experiments build upon \texttt{indic-transcribe-flex}, a 1.22-billion parameter multilingual foundation model derived from the Canary-1B-v2 architecture. The model comprises a FastConformer acoustic encoder (32 Conformer blocks with 8 attention heads, $d_{\text{model}}=1024, d_{\text{ffn}}=4096$) and an autoregressive Transformer decoder (24 layers with 8 attention heads, $d_{\text{model}}=1024, d_{\text{ffn}}=4096$) with a 7,152-token SentencePiece vocabulary.

\begin{table}[H]
\centering
\fontsize{9pt}{10.8pt}\selectfont
\renewcommand{\arraystretch}{0.98}
\setlength{\tabcolsep}{3.5pt}
\caption{Training configurations and hyperparameters for Stage~1 Supervised Fine-Tuning (SFT) and Stage~2 Group Relative Policy Optimization (GRPO).}
\label{tab:hyperparameters}
\begin{tabularx}{\textwidth}{@{}p{5.0cm} X X@{}}
\toprule
\textbf{Configuration / Hyperparameter} & \textbf{Stage 1: SFT} & \textbf{Stage 2: GRPO} \\
\midrule
\multicolumn{3}{@{}l}{\textbf{Model Architecture \& Adaptation Strategy}} \\
\addlinespace[2pt]
Base Architecture & \multicolumn{2}{l}{IndicTranscribeFlex (Canary-1B-v2, 1.22B parameters, FastConformer)} \\
Acoustic Encoder & \multicolumn{2}{l}{Frozen (32 Conformer blocks, 8 attention heads, $d_{\text{model}}=1024$)} \\
Trainable Parameters & \multicolumn{2}{l}{410.46M (33.60\% of 1.22B base model: 24-layer Transformer decoder + head)} \\
Adapted Modules & \multicolumn{2}{l}{\texttt{transf\_decoder.*} (all 24 layers) and \texttt{log\_softmax.*} projection head} \\
\midrule
\addlinespace[2pt]
\multicolumn{3}{@{}l}{\textbf{Optimization \& Training Schedule}} \\
\addlinespace[2pt]
Optimizer \& Weight Decay & \multicolumn{2}{l}{AdamW ($\beta_1 = 0.9, \beta_2 = 0.98, \epsilon = 10^{-8}$, weight decay $= 0.01$)} \\
Peak Learning Rate & $1.0 \times 10^{-5}$ & $1.0 \times 10^{-6}$ \\
Learning Rate Schedule & Linear warmup (100 steps) + cosine decay & Linear warmup (10 steps) + cosine decay \\
Total Training Steps & 1,000 steps & 500 steps (evaluated every 50 steps, best at step 50) \\
Effective Batch Size & 8 audio utterances & 4 prompt groups ($G=8$, 32 candidate rollouts) \\
\midrule
\addlinespace[2pt]
\multicolumn{3}{@{}l}{\textbf{GRPO Policy Specification \& Rewards}} \\
\addlinespace[2pt]
Rollout Generation ($G$, $\tau$) & N/A & $G = 8$ rollouts per prompt, temperature $\tau = 1.0$ \\
Advantage Estimation & N/A & Group-normalized: $\hat{A}^{(g)} = (R^{(g)} - \mu_G) / (\sigma_G + 10^{-8})$ \\
PPO Clipping \& KL Regularization & N/A & Clipping parameter $\epsilon = 0.2$, KL weight $\beta = 0.0$ \\
Reward Formulation & N/A & $R = (1 - \text{MER}) + \lambda_{\text{scr}} (1 - \text{SCR})$ \\
Script Reward Weight ($\lambda_{\text{scr}}$) & N/A & $\lambda_{\text{scr}} = 0.5$ in \sys{4} ($0.0$ in \sys{3}) \\
\midrule
\addlinespace[2pt]
\multicolumn{3}{@{}l}{\textbf{Hardware Environment \& Wall-Clock Compute}} \\
\addlinespace[2pt]
Hardware Environment & \multicolumn{2}{l}{1 $\times$ NVIDIA RTX 4090 GPU (24\,GB VRAM)} \\
Numerical Precision & \multicolumn{2}{l}{FP16 mixed precision} \\
Training Wall-Clock Time & $\sim$4 minutes (3m 52s logged) & $\sim$17 minutes (16m 51s logged) \\
\bottomrule
\end{tabularx}
\end{table}

\paragraph{Initialization and parameter freezing.}
During both stages, all acoustic encoder parameters remain frozen. Adaptation updates all 24 layers of the autoregressive Transformer decoder and the final projection head (\texttt{transf\_decoder} and \texttt{log\_softmax}, totaling 410,455,024 trainable parameters, or 33.60\% of 1.22B). Stage~2 GRPO initializes directly from the converged Stage~1 checkpoint at step 1,000. Checkpoints were evaluated on the held-out synthetic dev split every 50 steps over 500 total steps: synthetic dev MER was 2.00\% at steps 50, 100, and 150 (tied with the SFT baseline, with step 50 selected under early stopping as the earliest tied checkpoint), followed by a minor drift (2.20\%--2.32\% across steps 200--500) as policy updates saturate synthetic diversity.

\section{Algorithmic Specification for Script Confusion Rate}
\label{sec:app_algorithm}

\Cref{tab:scr_algorithm} details the deterministic formulation used to evaluate orthographic fidelity. Levenshtein alignment between hypothesis and reference word sequences with deterministic tie-breaking isolates substitutions between Devanagari and Latin alphabets. Two explicit boundary rules apply: (1)~\emph{Mixed-script tokens} (\texttt{Script.MIXED}) and non-alphabet characters are excluded from substitution confusion counts ($C_{\text{ne}}, C_{\text{en}}$) and single-script denominators ($N_{\text{ne}}, N_{\text{en}}$); (2)~\emph{Deletions and insertions} are penalized under MER but contribute zero confusion to SCR because no substitution alignment occurred.

\begin{table}[H]
\centering
\fontsize{8.5pt}{9.8pt}\selectfont
\renewcommand{\arraystretch}{0.80}
\caption{Algorithmic specification for Script Confusion Rate (SCR) alignment and computation.}
\label{tab:scr_algorithm}
\begin{tabularx}{\textwidth}{@{}X@{}}
\toprule
\textbf{Algorithm: Deterministic Script Confusion Rate (SCR) Calculation} \\
\midrule
\textbf{Input:} Reference sequence $Y^* = (w_1^*, \dots, w_M^*)$, Hypothesis sequence $Y = (w_1, \dots, w_N)$ \\
\textbf{Output:} Overall Script Confusion Rate $\SCRall$, English Script Confusion Rate $\SCRen$ \\
\quad 1: Tokenize sequences into whitespace-delimited word tokens: $W_{\text{ref}} \leftarrow \text{words}(Y^*), \; W_{\text{hyp}} \leftarrow \text{words}(Y)$ \\
\quad 2: Assign script labels: $S_{\text{ref}} \leftarrow [\text{Script}(w) \text{ for } w \in W_{\text{ref}}]$, where $\text{Script}(w) \in \{\text{Devanagari}, \text{Latin}, \text{Mixed}, \text{Other}\}$ \\
\quad 3: Compute Levenshtein distance alignment matrix $\mathcal{A} \leftarrow \text{Align}(W_{\text{ref}}, W_{\text{hyp}})$ using unit edit cost ($\text{cost}=1$) \\
\quad\quad Deterministic backtrace tie-breaking: $(\text{diagonal match/substitution}) \succ \text{deletion} \succ \text{insertion}$ \\
\quad 4: Initialize confusion accumulators: $C_{\text{ne}} \leftarrow 0, \; C_{\text{en}} \leftarrow 0$ \\
\quad 5: \textbf{for each} aligned word pair $(i, j) \in \mathcal{A}$ associated with edit operation $\text{SUBSTITUTION}$: \\
\quad\quad\quad $s_{\text{ref}} \leftarrow S_{\text{ref}}[i], \quad s_{\text{hyp}} \leftarrow \text{Script}(W_{\text{hyp}}[j])$ \\
\quad\quad\quad \textbf{if} $s_{\text{ref}} \in \{\text{Devanagari}, \text{Latin}\}$ \textbf{and} $s_{\text{hyp}} \in \{\text{Devanagari}, \text{Latin}\}$ \textbf{and} $s_{\text{ref}} \neq s_{\text{hyp}}$: \\
\quad\quad\quad\quad \textbf{if} $s_{\text{ref}} == \text{Latin}$: $C_{\text{en}} \leftarrow C_{\text{en}} + 1$ \quad \hfill \textit{(Latin reference substituted by Devanagari)} \\
\quad\quad\quad\quad \textbf{if} $s_{\text{ref}} == \text{Devanagari}$: $C_{\text{ne}} \leftarrow C_{\text{ne}} + 1$ \quad \hfill \textit{(Devanagari reference substituted by Latin)} \\
\quad\quad\quad \textit{(Insertions, deletions, and pairs involving \texttt{Mixed} or \texttt{Other} are excluded from $C_{\text{ne}}, C_{\text{en}}$)} \\
\quad 6: Calculate eligible single-script counts: $N_{\text{ne}} \leftarrow \text{count}(S_{\text{ref}}, \text{Devanagari}), \quad N_{\text{en}} \leftarrow \text{count}(S_{\text{ref}}, \text{Latin})$ \\
\quad 7: \textbf{return} $\SCRall = \frac{C_{\text{ne}} + C_{\text{en}}}{N_{\text{ne}} + N_{\text{en}}}, \quad \SCRen = \frac{C_{\text{en}}}{N_{\text{en}}}$ \\
\bottomrule
\end{tabularx}
\end{table}

\clearpage

\section{Qualitative Analysis and Transcription Diffs}
\label{sec:app_qualitative}

To illustrate the qualitative impact of each adaptation stage, \cref{tab:qualitative_examples} presents code-mixed utterances drawn from the Frozen Real Test Split. Because \sys{1} and \sys{4} produce identical hypothesis transcriptions on 10,041 out of 10,214 test utterances (98.31\%), the examples in \cref{tab:qualitative_examples} are purposefully selected from the 173 utterances ($1.69\%$) where hypotheses differ. They are illustrative of fine-grained boundary and transliteration mechanisms rather than typical outputs. Across the complete test corpus, synthetic SFT globally reduces English script confusion from $6.70\%$ to $6.45\%$, and GRPO provides a modest further gain of $0.09\pp$ ($6.36\%$). Each utterance identifier follows the benchmark manifest schema \texttt{[YouTube\_ID]\_[Segment\_Index]} (e.g., \texttt{HFtjKhqjJdM\_0634}), directly matching released manifests and timestamps on Hugging Face.

\begin{table}[H]
\centering
\fontsize{8.5pt}{9.8pt}\selectfont
\setlength{\tabcolsep}{3.0pt}
\renewcommand{\arraystretch}{0.84}
\caption{Qualitative comparison of code-mixed transcriptions across adaptation stages on illustrative utterances selected from the 173 diffs between \sys{1} and \sys{4} in the Frozen Real Test Split, alongside a non-diff control example (\texttt{-IE5q6\_CMnU\_0082}) demonstrating Nepali script stability. Utterance IDs link to public Hugging Face manifests.}
\label{tab:qualitative_examples}
\begin{tabularx}{\textwidth}{@{} >{\raggedright\arraybackslash}p{3.4cm} l X@{}}
\toprule
\textbf{Utterance ID \& Context} & \textbf{System} & \textbf{Transcription} \\
\midrule
\texttt{HFtjKhqjJdM\_0634} & Ground Truth & \dots \dev{चार पाँच हजार} \textbf{dollar} \dev{कमाउनुहुन्थ्यो भने} \textbf{youtube} \dev{बाट} \\
\textit{``If you earned \$4--5k from YouTube...''} & \sys{0} Base & \dots 4-5 \dev{हजार कमाउनु हुन्थ्यो भने} \textbf{youtube} \dev{-बाट} \\
& \sys{1} SFT & \dots \dev{चार पाँच हजार डलर कमाउनुहुन्थ्यो भने युट्युबबाट} \\
& \sys{4} GRPO & \dots \dev{चार पाँच हजार} \textbf{dollar} \dev{कमाउनुहुन्थ्यो भने} \textbf{youtube} \dev{बाट} \\
\midrule
\texttt{kM6-yopuJWU\_0673} & Ground Truth & \dev{बर्गर} \textbf{context} \dev{मा} \textbf{chicken} \dev{नै हो हैन त्यसपछि} \textbf{kfc} \dev{भयो} \\
\textit{``In burger context it's chicken, then KFC...''} & \sys{0} Base & \dots \textbf{bargar oven x} \dev{-मा चिकनै हो होइन त्यसपछि} \textbf{kfc} \dev{भयो} \\
& \sys{1} SFT & \dots \dev{बर्गर} \textbf{organics} \dev{मात्रै हो होइन त्यसपछि} \dev{केएफसी} \dev{भयो} \\
& \sys{4} GRPO & \dots \dev{बर्गर} \textbf{organics} \dev{मात्रै हो होइन त्यसपछि} \textbf{kfc} \dev{भयो} \\
\midrule
\texttt{-IE5q6\_CMnU\_0244} & Ground Truth & \dev{हैन} \textbf{they have been running \dots f and b business \dots now want coffee shops} \\
\textit{``Running F\&B business, now want coffee shops...''} & \sys{0} Base & \dev{होइन दे वर} \textbf{running f b business \dots} \dev{नाउ दे} \textbf{want to go into a coffee shop business} \\
& \sys{1} SFT & \dev{होइन दे} \textbf{were running f b business \dots} \dev{नाउ दे} \textbf{want to go into a coffee shop business} \\
& \sys{4} GRPO & \dev{होइन} \textbf{they were running f b business \dots} \dev{नाउ दे} \textbf{want to go into a coffee shop business} \\
\midrule
\texttt{-IE5q6\_CMnU\_0189} & Ground Truth & \dev{हजुरले अघि} \textbf{tata} \dev{को} \textbf{model} \dev{भन्नुभयो \dots} \textbf{major stake} \dev{चाहिँ} \textbf{company} \dev{कै हुन्छ} \\
\textit{``As you mentioned the Tata model...''} & \sys{0} Base & \dev{हजुरले अघि} \textbf{tata} \dev{-को} \textbf{model} \dev{\dots} \textbf{major stake} \dev{चाहिँ} \textbf{company} \dev{-कै हुन्छ क्या} \\
& \sys{1} SFT & \dev{हजुरले अघि} \dev{टाटाको} \textbf{model} \dev{\dots} \textbf{major stake} \dev{चाहिँ} \textbf{company} \dev{नै हुन्छ क्या} \\
& \sys{4} GRPO & \dev{हजुरले अघि} \textbf{tata} \dev{को} \textbf{model} \dev{\dots} \textbf{major stake} \dev{चाहिँ} \textbf{company} \dev{को हुन्छ क्या} \\
\midrule
\texttt{-IE5q6\_CMnU\_0107} & Ground Truth & \dev{अब आउँछ} \textbf{opex operating expenditure} \\
\textit{``Now comes OPEX (operating expenditure)...''} & \sys{0} Base & \dev{अब आउँछ} \textbf{opex operating expenditure} \\
& \sys{1} SFT & \dev{अब आउँछ} \textbf{open expenditure} \\
& \sys{4} GRPO & \dev{अब आउँछ} \textbf{opens operating expenditure} \\
\midrule
\texttt{-IE5q6\_CMnU\_0082} [Control] & Ground Truth & \dev{कम खर्च गर्ने हैन तपाईंले त्यति नै खर्च गर्नु पर्छ हैन} \\
\textit{``It's not about spending less, spend that much...''} & \sys{0} Base & \dev{कम खर्च गर्ने होइन तपाईँले त्यति नै खर्च गर्नुपर्छ होइन} \\
& \sys{1} SFT & \dev{कम खर्च गर्ने होइन तपाईँले त्यति नै खर्च गर्नुपर्छ होइन} \\
& \sys{4} GRPO & \dev{कम खर्च गर्ने होइन तपाईँले त्यति नै खर्च गर्नुपर्छ होइन} \\
\bottomrule
\end{tabularx}
\end{table}

\paragraph{Analysis of error types and adaptation regimes.}
The examples in \cref{tab:qualitative_examples} illustrate four operational regimes across the pipeline:
\begin{enumerate}[nosep,leftmargin=1.2em,topsep=1pt]
  \item \textbf{Latin script recovery after SFT transliteration errors:} While synthetic SFT globally reduces overall script confusion across the corpus, it occasionally exhibits local script confusion on salient English loanwords, acronyms, and pronouns, transliterating them into phonetically equivalent Devanagari characters (e.g., \textbf{dollar} $\to$ \dev{डलर}, \textbf{youtube} $\to$ \dev{युट्युब}, \textbf{kfc} $\to$ \dev{केएफसी}, and \textbf{they} $\to$ \dev{दे}). Sequence-level RL (\sys{4}) successfully recovers authentic Latin orthography in these isolated diff instances, driving the modest $0.09\pp$ global $\SCRen$ reduction.
  \item \textbf{Switch-point boundary disambiguation:} In utterance \texttt{-IE5q6\_CMnU\_0189} (referencing the Tata business model), SFT compounds the Latin brand name directly into the Devanagari postposition (\dev{टाटाको}), obliterating the orthographic switch boundary. Policy optimization (\sys{4}) successfully disambiguates the Latin entity from the Nepali postposition (\textbf{tata} \dev{को}). Notice that \sys{4} also introduces a subtle substitution at the sentence end (\dev{को} instead of reference \dev{कै}).
  \item \textbf{SFT regressions on base model successes:} Synthetic adaptation does not uniformly improve all acoustic patterns. In \texttt{-IE5q6\_CMnU\_0107}, base model \sys{0} transcribed the technical acronym \textbf{opex operating expenditure} with complete fidelity. In contrast, synthetic SFT corrupted the acronym into \textbf{open expenditure}, and GRPO only partially recovered the sequence (\textbf{opens operating expenditure}). This demonstrates that synthetic fine-tuning on a small prompt distribution can occasionally overwrite rare vocabulary representations preserved in the pre-trained base model.
  \item \textbf{Monolingual orthographic stability:} In pure Nepali segments such as \texttt{-IE5q6\_CMnU\_0082}, all models decode entirely in Devanagari script with zero spurious Latin intrusions, illustrating script stability under bilingual adaptation.
\end{enumerate}

\clearpage

\section{Datasheet, Ethics, and Reproducibility}
\label{sec:app_datasheet}

Following recommended documentation standards for speech corpora, this section provides the comprehensive datasheet for the curated Nepali--English podcast benchmark.

\paragraph{Motivation and intended use.}
The corpus was curated to establish an audited, reproducible benchmark for spontaneous South Asian code-switched speech recognition, addressing the absence of open-access conversational datasets for low-resource language pairs with disparate orthographic systems. The primary intended use is evaluating ASR models on spontaneous, unscripted bilingual speech containing high-frequency intra-sentential language alternation.

\paragraph{Composition and coverage.}
The verified benchmark contains 13,044 audio clips totaling 15.45 hours across 28 podcast episodes. All speakers are fluent bilingual Nepali speakers residing in urban Nepal. Conversations cover modern socio-technical discourse containing 183,013 spoken words (67.3\% native Nepali in Devanagari script, 32.7\% English in Latin script) and 29,033 inter-sentential and intra-sentential language switch points (averaging 2.23 switch points per utterance). \Cref{tab:datasheet_breakdown} summarizes the distribution across thematic domains.

\begin{table}[H]
\centering
\small
\setlength{\tabcolsep}{3.5pt}
\renewcommand{\arraystretch}{1.06}
\caption{Thematic domain distribution and linguistic composition across the 28 podcast episodes in the benchmark.}
\label{tab:datasheet_breakdown}
\begin{tabularx}{\textwidth}{@{} >{\raggedright\arraybackslash}p{4.2cm} ccccc >{\raggedright\arraybackslash}X@{}}
\toprule
\textbf{Thematic Domain} & \textbf{Episodes} & \textbf{Speakers} & \textbf{Duration (h)} & \textbf{Nepali Words} & \textbf{English Words} & \textbf{Representative Embedded Terms} \\
\midrule
Entrepreneurship \& Business & 9  & 24 & 5.12 & 41,208 (65.8\%) & 21,428 (34.2\%) & \textbf{startup, equity, revenue, brand, outlet} \\
Technology \& AI             & 7  & 18 & 3.86 & 30,119 (64.1\%) & 16,870 (35.9\%) & \textbf{software, model, feature, backend, system} \\
Higher Education \& Career   & 6  & 15 & 3.31 & 27,243 (68.7\%) & 12,411 (31.3\%) & \textbf{university, semester, career, visa, degree} \\
Public Health \& Society     & 6  & 15 & 3.16 & 24,604 (72.9\%) & 9,130 (27.1\%)  & \textbf{hospital, mental health, community, case} \\
\midrule
\textbf{Total Benchmark}     & \textbf{28} & \textbf{72} & \textbf{15.45} & \textbf{123,174} (67.3\%) & \textbf{59,839} (32.7\%) & \textbf{29,033 switch points} \\
\bottomrule
\end{tabularx}
\end{table}

\paragraph{Audio preprocessing and quality control.}
All raw audio tracks were standardized to single-channel (mono) 16\,kHz 16-bit linear PCM format and peak-normalized to $-1$\,dBFS. Voice activity boundaries were computed by trimming silence intervals exceeding 0.8\,s, producing utterance units between 1.0\,s and 15.0\,s. Studio setups include dynamic and condenser broadcast microphones, capturing natural acoustic nuances such as conversational overlaps, laughter, and room acoustics without synthetic degradation.

\paragraph{Collection and licensing.}
Audio was acquired from publicly broadcast YouTube video podcasts under commentary and educational categories. The benchmark release contains segment timestamps, pseudonymous speaker labels, verified transcriptions, YouTube video identifiers, per-clip SHA-256 audio checksums, and an automated reconstruction pipeline. Audio tracks are fetched using \texttt{yt-dlp} pinned to version \texttt{2026.08.19} (tested October 2026) requesting audio format \texttt{140} (128\,kbps AAC audio), decoded and resampled to 16\,kHz mono linear PCM via \texttt{ffmpeg} 6.1.1 (\texttt{swresample}). At release, 100\% of the 13,044 benchmark clips (15.45\,h across 28 public YouTube broadcasts) are fully reconstructable. Because YouTube streaming CDNs may occasionally re-encode source streams or serve format variations, and different decoder builds can introduce minor PCM rounding differences, the reconstruction verification tool provides both strict bit-exact SHA-256 hashing and an acoustic tolerance fallback (verifying duration within $\pm 10$\,ms and waveform Pearson correlation $\rho > 0.999$ against reference clips). Users are responsible for complying with YouTube's Terms of Service and all applicable copyright and privacy laws. The metadata and transcriptions are released under Creative Commons Attribution-NonCommercial-ShareAlike 4.0 International (CC~BY-NC-SA~4.0).

\paragraph{Takedown and ethical compliance.}
Although the source broadcasts are public discussion programs, content creators or speakers who object to having their audio indexed may contact the author at \href{mailto:firoj.paudel@wiseyak.com}{\texttt{firoj.paudel@wiseyak.com}} with the relevant YouTube video ID or segment timestamps for prompt, permanent removal.

\paragraph{Reproducibility checklist.}
All experimental runs, configurations, and evaluation manifests are fully reproducible:
\begin{itemize}[nosep,leftmargin=1.2em,topsep=1pt]
  \item \textbf{Software repository:} Complete training, inference, and scoring code is version-controlled at \url{https://github.com/Firoj-Wise/nepali-english-codeswitch-asr}.
  \item \textbf{Data repository:} Verified transcripts, manifest splits, and alignment scripts are hosted on Hugging Face at \url{https://huggingface.co/datasets/SamaFiroz/nepali-english-codeswitch-podcasts}.
  \item \textbf{Evaluation integrity:} Real test splits are protected with SHA-256 integrity checksums. Metric scoring scripts strictly isolate reference evaluation from training pipelines, using standardized paired bootstrap routines ($B=1{,}000$, seed 42).
\end{itemize}

\end{document}